# UAV ROUTE PLANNING FOR MAXIMUM TARGET COVERAGE

Murat Karakaya [1]

[1]Department of Computer Engineering, Atilim University, Ankara, Turkey

## ABSTRACT

*Utilization of Unmanned Aerial Vehicles (UAVs) in military and civil operations is getting popular. One of the challenges in effectively tasking these expensive vehicles is planning the flight routes to monitor the targets. In this work, we aim to develop an algorithm which produces routing plans for a limited number of UAVs to cover maximum number of targets considering their flight range.*

*The proposed solution for this practical optimization problem is designed by modifying the Max-Min Ant System (MMAS) algorithm. To evaluate the success of the proposed method, an alternative approach, based on the Nearest Neighbour (NN) heuristic, has been developed as well. The results showed the success of the proposed MMAS method by increasing the number of covered targets compared to the solution based on the NN heuristic.*

## 1. INTRODUCTION

This document describes, and is written to conform to, author guidelines for the journals of AIRCC series. It is prepared in Microsoft Word as a .doc document. Although other means of preparation are acceptable, final, camera-ready versions must conform to this layout. Microsoft Word terminology is used where appropriate in this document. Although formatting instructions may often appear daunting, the simplest approach is to use this template and insert headings and text into it as appropriate. The importance and the impact of using Unmanned Aerial Vehicles (UAVs) in military and civil operations are increasing [3,4,5]. One of the issues faced for efficient usage of UAVs is planning the flight routes to monitor all or the maximum number of the given targets. This problem is related with the Multiple Travelling Salesman Problem (mTSP) [1] and the Vehicle Routing Problem (VRP) [6]. In these well-defined problems, it is mostly assumed that travelling salesmen or vehicles should visit all the targets and the target function is defined as to find a minimum-distant route. Even, in the constraint versions of the mTSP and VRP, some other restrictions (visiting time windows, number of depots, etc.) are included; it is still assumed that there exists enough number of travelling salesmen or vehicles to cover all the given locations.

However, in reality the number and flight range of UAVs might be insufficient to cover all the targets. As a result, the maximization of the number of targets covered by the limited number of UAVs can be defined as a new problem. Thus, this article presents a solution for this practical optimization problem by modifying the Max-Min Ant System (MMAS) algorithm [2] accordingly.

In the proposed solution, each ant constructs routes for the given number of UAVs using pheromone and heuristic information. After each iteration, the solution which covers more targets with less route distance is selected as the iteration-best solution and the pheromone values of the edges on that route are increased. According to the termination condition, the algorithm stops and outputs the best route found so far as the result. To evaluate the success of the proposed method, another approach, based on the Nearest Neighbour (NN) heuristic, is designed as well. In this solution, an UAV always select the nearest target to move on until its remaining flight range urges the UAV to return the base.

Both solutions are implemented using MASON simulation library [7] and compared by extensive experiments with different parameters and standard TSP data files [9]. The results showed the success of the proposed MMAS method by increasing the number of covered targets up to 10% compared to the solution based on the NN heuristic.

## 2. PROBLEM DEFINITION

We assume that we are given the location of each target along with the base location, the number, and flight range of the UAVs. The problem is to create routes for each UAV such that any target is visited by only one UAV and once, every UAV's route distance has to be equal or less than the flight range, and the number of total targets visited by the all UAVs is maximized. Thus the target function is to maximize the number of targets to be visited by the all UAVs. The constraints are the flight range and the number of UAVs.

## 3. MAX-MIN ANT SYSTEM

Stützle and Hoos proposed the Max-Min Ant Colony System (MMAS) as a successful alternative to Ant System (AS) [8]. In the referenced work, they show the relative success. The basic difference between the MMAS and AS is the setting up limits on the maximum and minimum values of the pheromone values that can be compiled on an edge. We apply MMAS to find a route planning to cover most of the targets as explained below.

## 4. APPLYING MMAS TO TARGET COVERAGE PROBLEM

Below, we first explain the MMAS basics and then provide the algorithm to generate a solution to cover maximum number of targets.

### 2.1. Selecting Next Target

In MMAS, each artificial ant tries to create a route planning for all the UAVs by visiting targets considering the given problem constraints. Beginning from the base, each ant calculates the probability of movement from the current location to the all unvisited targets as in the following formula:

$$p_{ij} = \frac{(\tau_{ij})(\eta_{ij})^{\beta}}{\sum_{i,j \notin M_k}(\tau_{ij})(\eta_{ij})^{\beta}}, \qquad i,j \notin M_k \qquad (1)$$

In Eq (1), $i$ is the current location, $j$ is the possible next location, $\tau_{ij}$ is the pheromone value between two locations, $\eta_{ij}$ is the heuristic value between two locations, $\beta$ is the coefficient for the heuristic parameter, and $M_k$ is the memory for storing list of the targets which either are already visited by the ant $k$ or cannot be accessible with the remaining flight range. Thus, $P_{ij}$ is the normalized probability of ant $k$ to move from target $i$ to target $j$. After calculating the movement probability for all the targets, a random number between 0 and 1 is generated to select the next target according to total probabilities of all the possible targets. If all the targets' probability is 0, it means that either all the targets are visited or the flight range is not enough to visit any targets any more. Then, ant returns to the base. Thus, a route for a UAV is completed. The ant begins a new route for the next UAV with a refreshed flight range. When all the routes are prepared for all the UAVs an iteration of the ants has been finished. Each ant builds its own route planning simultaneously by exploiting the experiences of other ants by sensing the pheromone values in the formula.

## 4.2. Assigning Initial Pheromone Values

The initial pheromone values ($\tau_0$) between all target in the set ($H$) are initiated to the selected maximum value ($\tau_{max}$). We calculated the $\tau_{max}$ as in Eq.(4).

$$\tau_0 = \tau_{max} \tag{2}$$

$$\tau_{ij} = \tau_0, \quad i,j \in H \tag{3}$$

$$\tau_{max} = \frac{1}{p * c_{init}} \tag{4}$$

In Eq. (4), $p$ is the evaporation parameter and $c_{init}$ is the cost of the initial solution. Cost of a solution is calculated as follows:

$$c = 1 - \frac{T_{visited}}{T_{all}} \tag{5}$$

Thus, if a routing plan can lead to visit all the targets, its cost will be zero. The initial solution is constructed using Nearest Neighbors heuristic.

The minimum pheromone value is defined as

$$\tau_{min} = (1-p)^{\frac{iteration}{10}} * \tau_{max} \tag{6}$$

As a result of Eq. (6), any edge would have pheromone at least ten times evaporated value of the maximum pheromone value. Thus, we do not allow unvisited edges get very low pheromone values which otherwise would decrease their probability.

## 4.3. Updating Pheromone Values

After completing a tour, each ant calculates the tour cost as given in Eq. (5). Before applying any pheromone update on the targets, first evaporation should take place. Thus, all the pheromone values between all the targets are decreased using the evaporation parameter value ($p$) as in the following formula:

$$\tau_{ij} = (1 - p)\tau_{ij}, \text{ i,j} \in \text{H} \tag{7}$$

Then, all targets on the route constructed by the ant ($R_k$) receive an update depending on the cost of the tour ($c$):

$$\tau_{ij} = \tau_{ij} + (\frac{1}{c}) \text{ , } i,j \in R_k \tag{8}$$

Eq. (8) dictates that the solutions with less cost, that is covering more targets, leave more pheromone on the paths to provide positive feedback for the other ants.

### 4.4. Calculating Heuristic Value

The heuristic value ($\eta_{ij}$) between two locations is defined as $\eta_{ij} = \dfrac{1}{d_{ij}}$, where $d_{ij}$ is the distance between the locations.

### 4.5. Algorithm

Using the steps defined above an implementation of the MMAS is given in Table 1. We input the target list (H), the distances between the targets ($d_{ij}$), the flight range (*FR*), and the number of UAVs (*UAV*) to the algorithm. The algorithm first calculates an initial solution by applying the NN heuristic. By using the cost of the initial solution, minimum and maximum pheromone values are set. Then, using the distance matrix, the heuristic values are calculated. After creating a number of ants (*m*), each ant builds its solution and updates the pheromone values according to the cost of the solution. When, a pre-defined number of iterations has been executed algorithm terminates by outputting the best solution found so far.

Table 1. Pseudocode for the proposed algorithm.

```
MaxTarget (H, dᵢⱼ , FR, UAV)
 {
    UAV_used = 0;
    remaining_Range = FR;
    Rₙₙ= NN(H, Mᵢⱼ , FR );
    init_Pheromone_Values();
    init_Heuristic(Mᵢⱼ);
    create_Ants(m, base);

    while (!end_condition_satisfied)
    {
        for each ant
        {
           while (UAV_used<UAV)
           {
                next = find_Next_Target();
                if (base_Reachable(next))
                {
                   move(next);
                   remaining_Range -= d_current,next;
                   target_Number ++;
                 }
                else
                {
                   move(base);
                   UAV_used++;
                   remaining_Range = FR;
                  }
            } //end_while
           evoporate_Pheromone();
           update_Pheromone();
           update_Best_Solution();
        }// end_for_each_ant
    }
    return (Best_Solution);
}
```

## 3. SIMULATION MODEL AND RESULTS

We have implemented the proposed MMAS solution using the MASON simulation library (Luke et al. 2003). The simulation and MMAS parameters with the default values are given in Table 2. We use several different TSP data files (TSPLIB, 1995) with various flight range and UAV number to observe the results. Below we report only the preliminary results for the TSP data file name CH150. The first location in the data file is selected as the base where all the UAVs are assumed to be located at the beginning and must return to it at the end of the flight. Thus, the total number of the targets is 149.

Table 2.  Simulation Parameters and default values.

| Parameter | Definition | Default Value |
|---|---|---|
| DF | Data File | CH150.dat |
| T | Total Number of Targets | 149 |
| $d_{ij}$ | Distance matrix | Calculated according to the input file. |
| FR | Flight range | 3 |
| $p$ | Pheromone evaporation | 0.01 |
| $\beta$ | Heuristic effect factor | 7 |
| $m$ | Number of ants | 151 |
| $t$ | Iteration number | 1000 |

To determine the flight range (*FR*) we define a parameter called Critical Distance (*CD*). The *CD* is the distance of the farthest target from the selected base. We test three *FR*s with respect to the *CD* as Case 1: FR = CD, Case 2: FR = CD/2, and Case 3: FR =CD*2.

The main performance metric, Target Coverage (*TC*), is the ratio of the number of the targets visited by all the UAVs to the existing targets as formulated below:

$$TC = \frac{T_{visited}}{T_{all}} *100 \tag{9}$$

To obtain the results, we run each simulation 10 times and get the averages of these results to find the mean values.

## 5.1. Results for Case 1

In the first test, FR is set CD and the results are presented in Table 3. The first column shows the number of UAVs, the second and the third give the TC results for the NN and MMAS heuristics respectively.

Table 3.  The target coverage ratios for the heuristics when FR = CD.

| UAV | $TC_{NN}$ | $TC_{MMAS}$ |
|---|---|---|
| 1 | 22% | 28% |
| 3 | 45% | 63% |
| 5 | 70% | 85% |
| 6 | 85% | 95% |
| 7 | 90% | 99% |
| 9 | 92% | 100% |
| 11 | 97% | 100% |
| 13 | 100% | 100% |

As seen in Table 3, the MMAS can generate more target coverage compared to the NN heuristic. For example, to cover all the targets, the MMAS can need only 9 UAVs whereas NN requires 13. This results shows that the MMAS can use UAVs much more efficiently to cover most of the targets with respect to the NN heuristic.

## 5.2. Results for Case 2

When the FR is decreased to half of the CD value, the generated results are as in Table 4. As the FR is not enough to reach some of the targets, increasing the number of UAVs cannot help after some point. In the experiments, we observe that the MMAS can access all possible targets with using 11 UAVs where the NN needs 14 to cover the same number of targets.

Table 4. The target coverage ratios for the heuristics when FR = CD/2.

| UAV | $TC_{NN}$ | $TC_{MMAS}$ |
|---|---|---|
| 1 | 11% | 12% |
| 3 | 20% | 29% |
| 5 | 30% | 35% |
| 7 | 34% | 38% |
| 9 | 36% | 40% |
| 11 | 38% | 41% |
| 13 | 40% | 41% |
| 14 | 41% | 41% |
| 1 | 11% | 12% |

## 5.3. Results for Case 3

For the last experiment, we set the FR double length of the CD. Since the FR is relatively large, we expect to cover all the targets with less number of UAVs. The MMAS performs better for any number of UAVs in this set of experiments as well. For instance, 4 UAVs are successfully routed by the MMAS to cover all the targets while the NN prepares a routing plan for the same number of UAVs missing 4% of the targets.

Table 5. The target coverage ratios for the heuristics when FR = CD*2.

| UAV | $TC_{NN}$ | $TC_{MMAS}$ |
|---|---|---|
| 1 | 11% | 12% |
| 3 | 20% | 29% |
| 5 | 30% | 35% |
| 7 | 34% | 38% |
| 9 | 36% | 40% |
| 11 | 38% | 41% |
| 13 | 40% | 41% |
| 14 | 41% | 41% |
| 1 | 11% | 12% |

## 3. CONCLUSIONS

In this work, we define a practical problem faced in route planning of a limited number of UAVs to cover maximum number of the given targets with a pre-defined flight range. We propose to adapt the MMAS meta-heuristic to solve this problem. After implementing the proposed solution we compare the results with an alternative heuristic, namely the Nearest Neighbor.

The preliminary results show the effectiveness of the MMAS in route planning. We would like to extend the work by defining different performance metrics and executing the experiments with different location set ups.

# REFERENCES

[1]     Bektas, T. (2006). The multiple traveling salesman problem: an overview of formulations and solution procedures. Omega, 34(3), 209-219.

[2]     Dorigo, M., Blum, C. (2005). Ant colony optimization theory: A survey. Theoretical computer science, 344(2), 243-278.

[3]     Ercan, C., Gencer, C. (2013). Dinamik nsansız Hava Sistemleri Rota Planlaması Literatür Ara tırması ve nsansız Hava Sistemleri Çalı ma Alanları. Pamukkale Üniversitesi Mühendislik Bilimleri Dergisi, 19(2), 104-111.

[4]     Everaerts, J. (2008). The use of unmanned aerial vehicles (UAVs) for remote sensing and mapping. The International Archives of the Photogrammetry, Remote Sensing and Spatial Information Sciences, 37, 1187-1192.

[5]     Glade, D. (2000). Unmanned aerial vehicles: Implications for military operations. Air Univ Press Maxwell Afb Al.

[6]     Laporte, G. (1992). The vehicle routing problem: An overview of exact and approximate algorithms. European Journal of Operational Research, 59(3), 345-358.

[7]     Luke, S., Balan, G. C., Panait, L., Cioffi-Revilla, C., & Paus, S. (2003). MASON: A Java multi-agent simulation library. In Proceedings of Agent 2003 Conference on Challenges in Social Simulation (Vol. 9).

[8]     Stützle, T., & Hoos, H. H. (2000). MAX–MIN ant system. Future generation computer systems, 16(8), 889-914.

[9]     TSPLIB, 1995, TSPLIB, [WWW document; retrieved August 2013] URL http://comopt.ifi.uni-heidelberg.de/software/TSPLIB95/

**Author**

**Murat KARAKAYA** received the B.S.E.E. degree in 1991 from the Turkish Military Academy,   Ankara, Turkey, and the M.S. and Ph. D. degrees in Computer Engineering from the Bilkent University, Ankara, Turkey in 2000 and 2008, respectively. From 2000 to 2005, he worked as an instructor and software engineer at the Turkish Military Academy, Ankara, Turkey. From 2008 to 2012, he worked as an instructor and software engineer at the Turkish Military School of Electronics, Communications and Information Systems (MEBS) and Turkish Military Academy 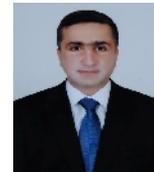 (KHO), Ankara, Turkey. He joined the faculty of Atılım University in 2012 and is currently an Asst. Professor in the department of Computer Engineering, Ankara, Turkey. His research interests are natural computing, sensor networks, peer-to-peer  networks, and communication protocol design.